\documentclass[conference]{IEEEtran}

\usepackage{cite}

\ifCLASSINFOpdf
\else
\fi

\usepackage{xcolor}

\usepackage[utf8x]{inputenc}
\usepackage{url}
\usepackage{array}
\usepackage{graphicx}
\usepackage{booktabs}
\usepackage{multirow}

\ifCLASSOPTIONcompsoc
\usepackage[caption=false,font=normalsize,labelfont=sf,textfont=sf]{subfig}
\else
\usepackage[caption=false,font=footnotesize]{subfig}
\fi

\begin{document}

\title{PELESent: Cross-domain polarity classification using distant supervision}

\author{\IEEEauthorblockN{Edilson A. Corr\^ea Jr, Vanessa Q. Marinho, Leandro B. dos Santos, \\ Thales F. C. Bertaglia, Marcos V. Treviso, Henrico B. Brum}
\IEEEauthorblockA{Institute of Mathematics and Computer Science\\\
University of S\~ao Paulo (USP)\\
S\~ao Carlos, S\~ao Paulo, Brazil\\
Email: {\tt \{edilsonacjr,vanessaqm,leandrobs,thales.bertaglia,marcostreviso,henrico.brum\}@usp.br}}}

\maketitle

\begin{abstract}
The enormous amount of texts published daily by Internet users has fostered the development of methods to analyze this content in several natural language processing areas, such as sentiment analysis. The main goal of this task is to classify the polarity of a message. Even though many approaches have been proposed for sentiment analysis, some of the most successful ones rely on the availability of large annotated corpus, which is an expensive and time-consuming process. In recent years, distant supervision has been used to obtain larger datasets. So, inspired by these techniques, in this paper we extend such approaches to incorporate popular graphic symbols used in electronic messages, the \emph{emojis}, in order to create a large sentiment corpus for Portuguese. Trained on almost one million tweets, several models were tested in both same domain and cross-domain corpora. Our methods obtained very competitive results in five annotated corpora from mixed domains (Twitter and product reviews), which proves the domain-independent property of such approach. In addition, our results suggest that the combination of emoticons and emojis is able to properly capture the sentiment of a message.
\end{abstract}

\IEEEpeerreviewmaketitle

\section{Introduction}

In the last few years, Sentiment Analysis has become a prominent field in natural language processing (NLP), mostly due to its direct application in several real-world scenarios~\cite{pang2008opinion}, such as product reviews, government intelligence, and the prediction of the stock markets. 
One of the main tasks in Sentiment Analysis is the polarity classification, i.e., classifying texts into categories according to the emotions expressed on them. In general, the classes are positive, negative and neutral. 

A popular application of polarity classification is in social media content. Microblogging and social networks websites, such as Twitter, have been used to express personal thoughts. According to Twitter's website\footnote{\url{https://business.twitter.com/en/basics.html}}, more than 500 million short messages, known as tweets, are posted each day. The analysis of this type of content is particularly challenging due to its specific language, which is mostly informal, with spelling errors, out of the vocabulary words, as well as the usage of emoticons and emojis to express ideas and sentiments. 

Machine learning methods have been widely applied to polarity classification tasks in the context of social networks. This is particularly evident in shared tasks such as the SemEval Sentiment Analysis tasks~\cite{rosenthal2015semeval,nakov2016semeval}, where these methods usually outperform lexical-based approaches. However, a major drawback of machine learning is its high dependency on large annotated corpora, and since manual annotation usually is time-consuming and expensive~\cite{pan2010cross}, many non-English languages lack this type of resource or, when existing, are very limited and specific, as it is the case for Portuguese.



In this paper, we adapt a distant supervision approach~\cite{go2009twitter} to annotate a large number of tweets in Portuguese and use them to train state-of-the-art methods for polarity classification. We applied these methods in manually annotated corpora from the same domain (Twitter) and cross-domain (product reviews). The obtained results indicate that the proposed approach is well suited for both: same domain and cross-domain. Moreover, it is a powerful alternative to produce sentiment analysis corpora with less effort than manual annotation.

This paper is organized as follows. Section~\ref{sec:related} gives a brief overview of some approaches for sentiment analysis and presents some works that have applied distant supervision to this task. Our approach is described in Section~\ref{sec:methodology}. The evaluation corpora, machine learning algorithms and results are given in Section~\ref{sec:experiment}. Finally, our conclusions are drawn in Section~\ref{sec:conclusion}.

\section{Related Work}\label{sec:related}

Currently, methods devised to perform sentiment analysis and, more specifically, polarity classification range from machine learning to lexical-based approaches. While machine learning methods have proved useful in scenarios where a large amount of training data is available along with top quality NLP resources (such as taggers, parsers and others), they usually have low performance in opposite scenarios. Since most non-English languages face resource limitations, for example Portuguese, lexical-based approaches have become very popular. Some works following this line are~\cite{souza2011construction,BalageFilho2013,avancco2014lexicon}. 

Another alternative for languages with fewer resources is the use of hybrid systems, which combine machine learning and lexical-based methods. Avan{\c{c}}o et al.\cite{avanccoimproving} showed that this combination outperforms both individual approaches. This may imply that the development of better individual elements will lead to better results in the final combination.

Machine learning approaches rely on document representations, normally vectorial ones with features like $n$-grams~\cite{pang2008opinion}, a simple example is the bag-of-words model. Once a representation has been chosen, several classification methods are available, such as Support Vector Machines~(SVM), Naive Bayes~(NB), Maximum Entropy~(MaxEnt), Conditional Random Fields~(CRF), and ensembles of classifiers~\cite{nakov2016semeval}.

Apart from the traditional features, such as $n$-grams, some researchers have taken advantage of word embeddings, which are known to capture some linguistic properties, such as semantic and syntactic features. A well-known example of word embeddings is \emph{Word2Vec}~\cite{mikolov2013efficient,mikolov2013distributed}. Algebraic operations, such as sum or average, can be applied to convert word vectors into a sentence or document vector~\cite{zhou2016ecnu,correa2017nilc}. However, this representation does not consider the order of the words in the sentence. 

Paragraph vectors~\cite{mikolovdoc2vec} (also known as \emph{Doc2Vec}) can be understood as a generalization of \emph{Word2Vec} for larger blocks of text, such as paragraphs or documents. This technique has obtained state-of-the-art results on sentiment analysis for two datasets of movie reviews~\cite{mikolovdoc2vec}. The main goal of these dense representations is to predict the words in those blocks. Two models were proposed by Le and Mikolov~\cite{mikolovdoc2vec}, in which one of them accounts for the word order.

In addition, deep neural networks also consider the word order. Their methods have achieved good results in sentiment analysis, as shown in~\cite{socher2013recursive,kalchbrenner2014convolutional,kim2014cnn} and in the SemEval Sentiment Analysis Tasks~\cite{rosenthal2015semeval,nakov2016semeval}. Nevertheless, these approaches need large datasets for training. Distant Supervision is a good alternative to obtain these datasets for the training/pre-training of deep neural networks~\cite{kalchbrenner2014convolutional,severyn2015unitn,deriu2016swisscheese}.

Distant supervision is an alternative to create large datasets without the need of manual annotation. Some works have reported the use of emoticons as semantic indicator for sentiment~\cite{read2005using,go2009twitter,pak2010twitter,severyn2015unitn}, while others use emoticons and hashtags for the same purpose~\cite{davidov2010enhanced,kouloumpis2011twitter}. Go et al.\cite{go2009twitter}, the first work to apply distant supervision to Twitter data, collected approximately 1.6 million of tweets containing positive and negative emoticons~--~e.g. ``:)'' and ``:(''~-- equally distributed into two classes. They combined sets of features~--~unigrams, bigrams, part-of-speech (POS) tags~-- in order to train machine learning algorithms~(NB, MaxEnt and SVM) and evaluate those in manually annotated datasets. The best accuracy was achieved using unigram and bigram as features for a MaxEnt classifier.

Severyn and Moschitti\cite{severyn2015unitn} used Distant Supervision to pre-train a Convolutional Neural Network~(CNN). An architecture similar to the one proposed by Kim~\cite{kim2014cnn}. The network is composed of a first layer to convert words in dense vectors, following a single convolutional layer with a non-linear activation function, max pooling and soft-max. Deriu et al.\cite{deriu2016swisscheese} used a combination of 2 CNNs with a Random Forest classifier. However, this approach did not obtain improvements with distant supervision.

Despite the numerous studies and investigations of different techniques and methods for polarity classification, the problem of relying on large annotated corpora remains open and the difficulty is intensified in non-English languages. In this paper, our contributions are the adapted framework for building polarity classification corpus to Portuguese, the built corpus itself and an evaluation on different state-of-the-art methods using this corpus, for same domain and cross-domain corpora.


\section{Approach}\label{sec:methodology}

Following the approach of Go et al.~\cite{go2009twitter}, we initially collected a large amount of tweets in order to create the distant supervision corpus. Only tweets in Portuguese were crawled, and no specific queries were employed. In total, 41 million of tweets were collected.

\begin{figure}[h]
\centering
\subfloat[\label{}]{\includegraphics[width=0.08\linewidth]{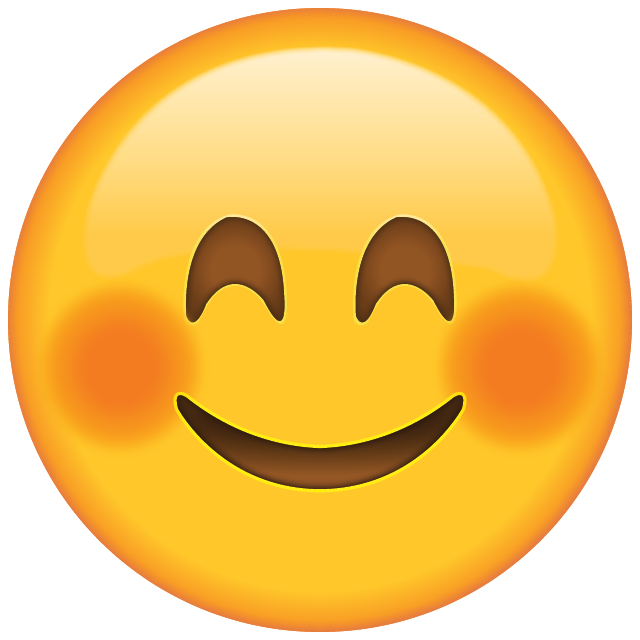}}
\subfloat[\label{}]{\includegraphics[width=0.08\linewidth]{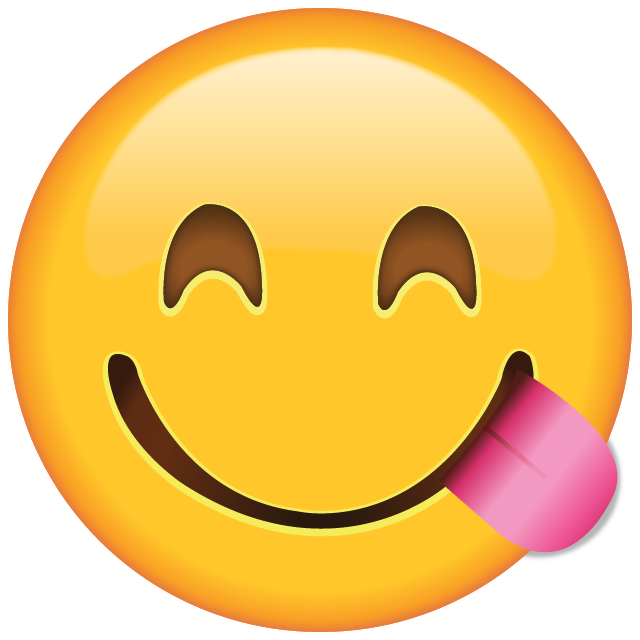}}
\subfloat[\label{}]{\includegraphics[width=0.08\linewidth]{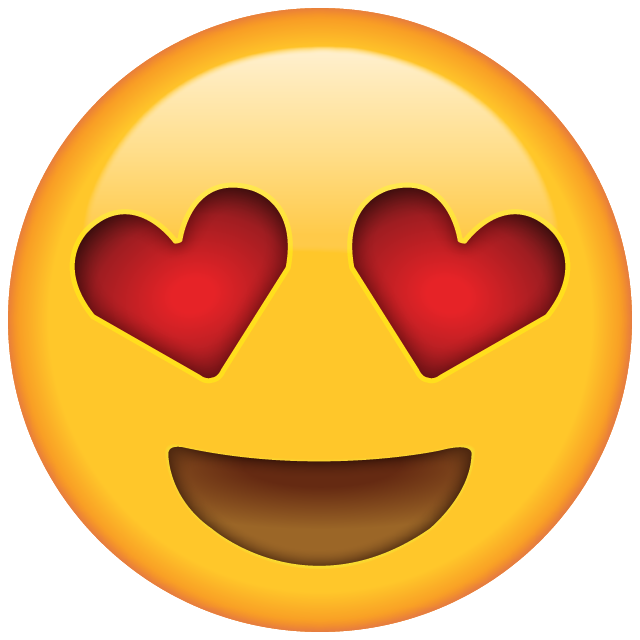}}
\subfloat[\label{}]{\includegraphics[width=0.08\linewidth]{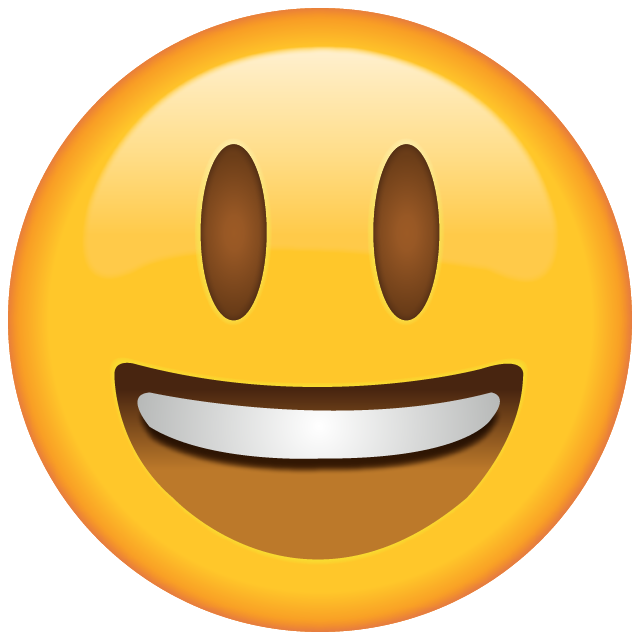}}
\subfloat[\label{}]{\includegraphics[width=0.08\linewidth]{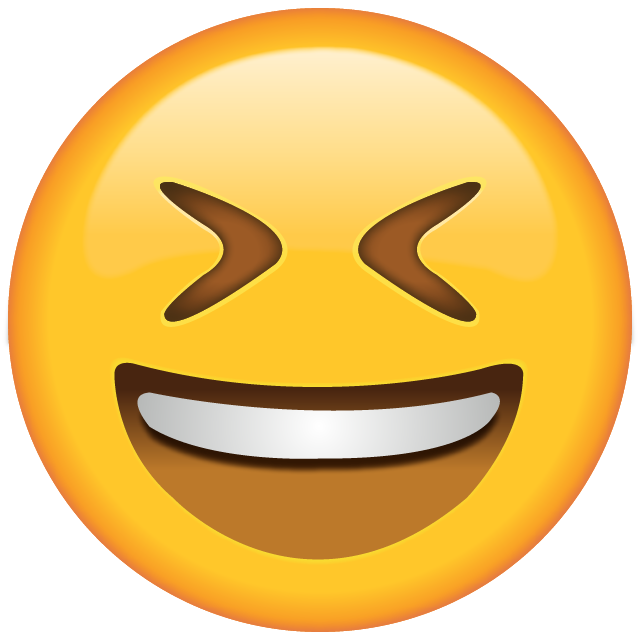}}
\subfloat[ \label{}]{\includegraphics[width=0.08\linewidth]{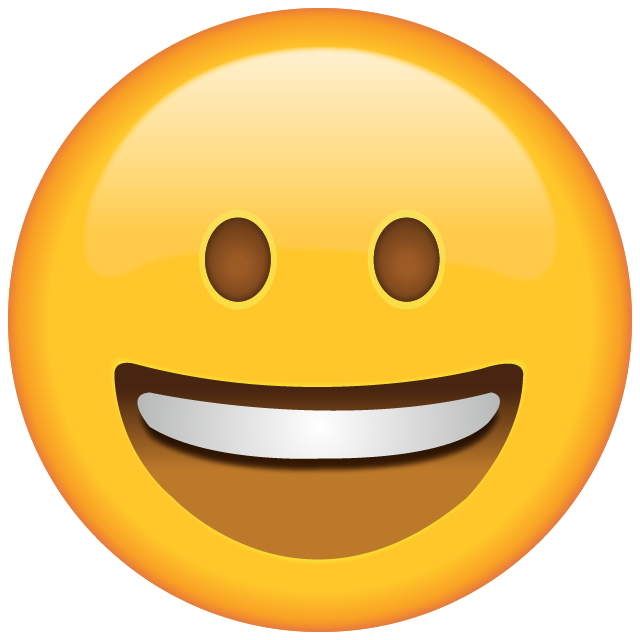}}
\subfloat[ \label{}]{\includegraphics[width=0.08\linewidth]{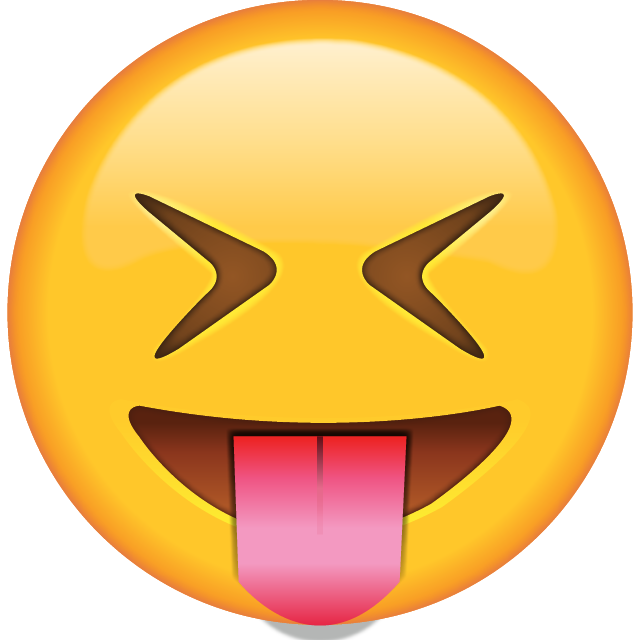}}
\caption{All emojis used to represent positive emotion. Their respective unicodes are: (a) U+1F60A, (b) U+1F60B, (c) U+1F60D, (d) U+1F603, (e) U+1F606, (f) U+1F600, and (g) U+1F61D.}\label{fig:positive_emojis}
\end{figure}

\begin{figure}[h]
\centering
\subfloat[\label{}]{\includegraphics[width=0.08\linewidth]{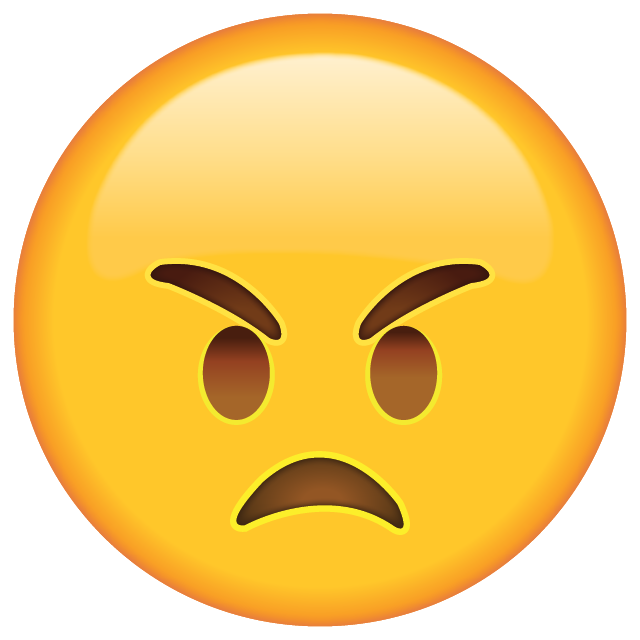}}
\subfloat[\label{}]{\includegraphics[width=0.08\linewidth]{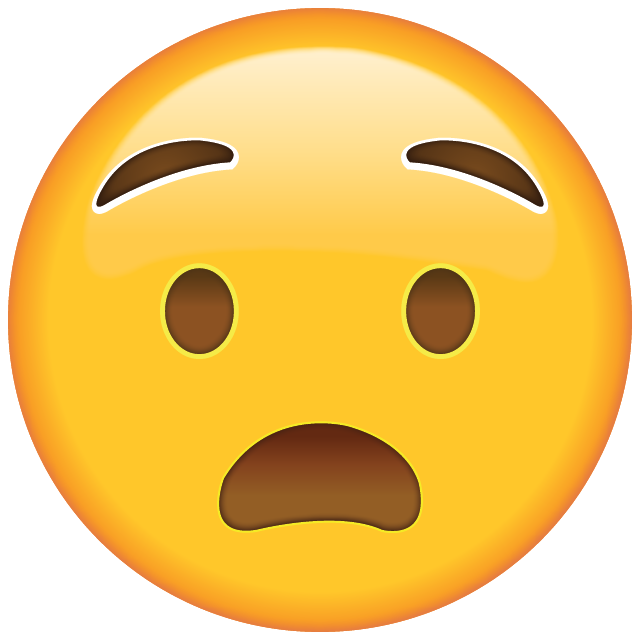}}
\subfloat[\label{}]{\includegraphics[width=0.08\linewidth]{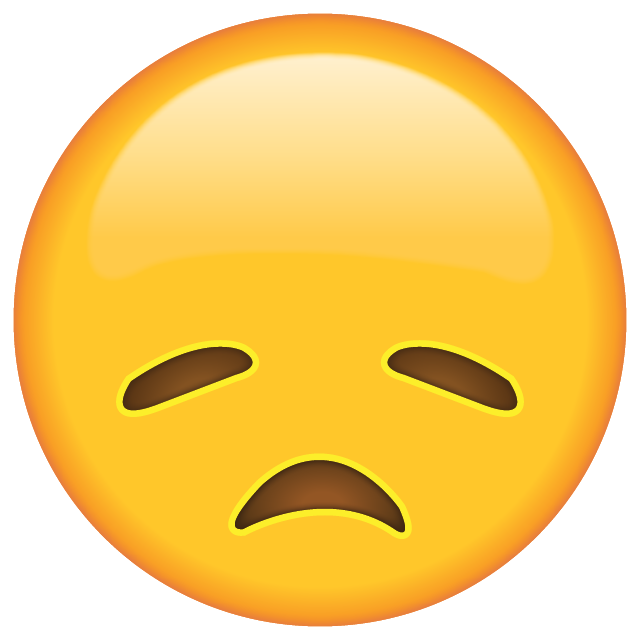}}
\subfloat[\label{}]{\includegraphics[width=0.08\linewidth]{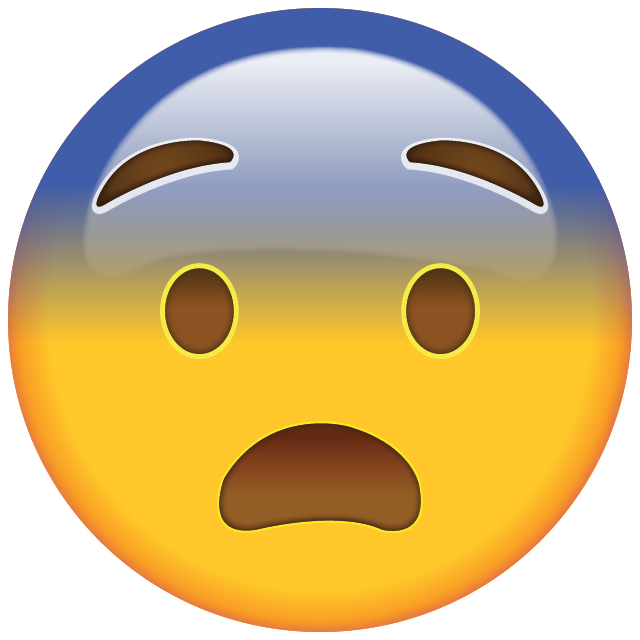}}
\subfloat[\label{}]{\includegraphics[width=0.08\linewidth]{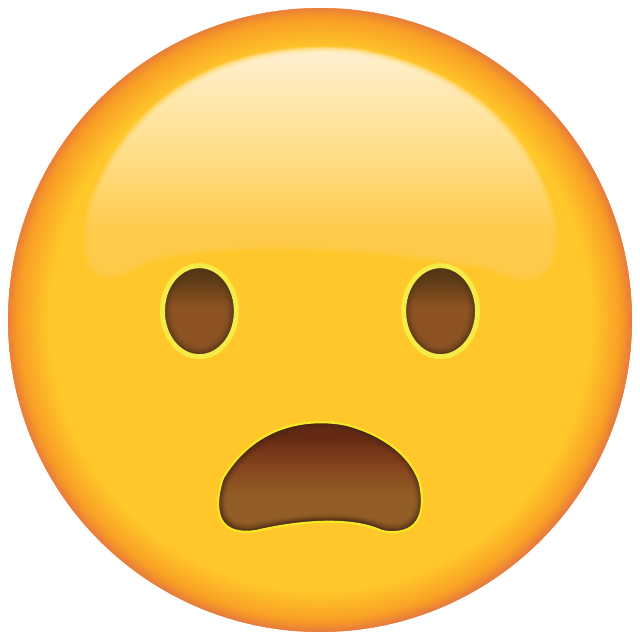}}
\subfloat[\label{}]{\includegraphics[width=0.08\linewidth]{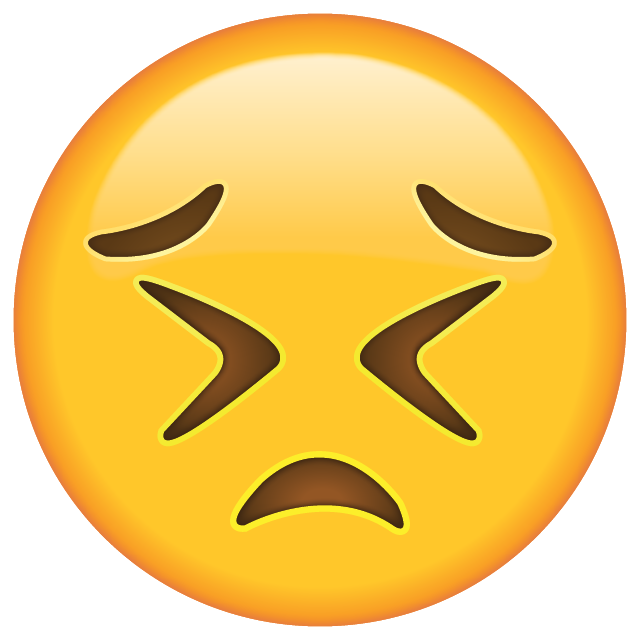}}
\\
\subfloat[\label{}]{\includegraphics[width=0.08\linewidth]{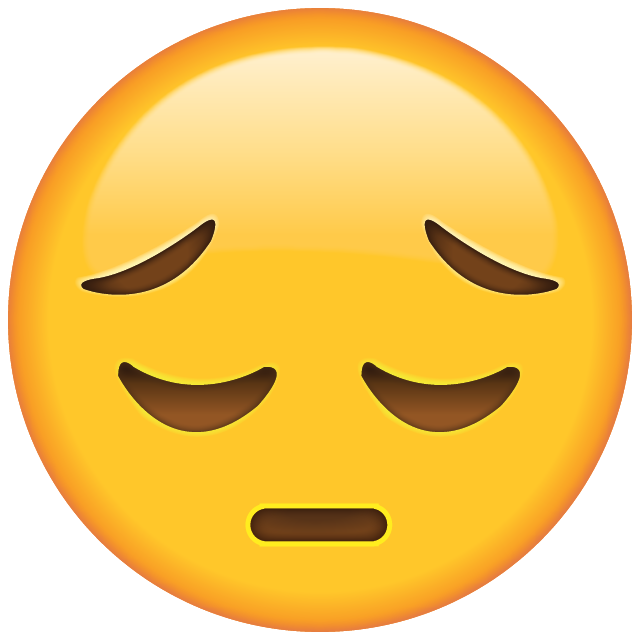}}
\subfloat[\label{}]{\includegraphics[width=0.08\linewidth]{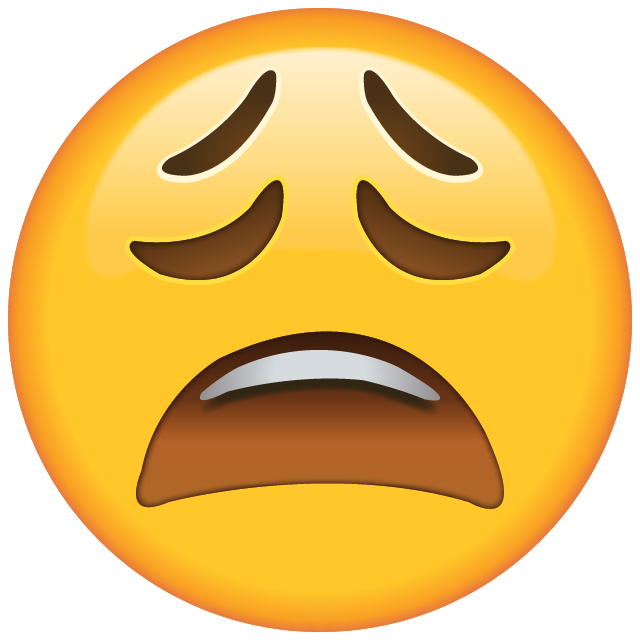}}
\subfloat[\label{}]{\includegraphics[width=0.08\linewidth]{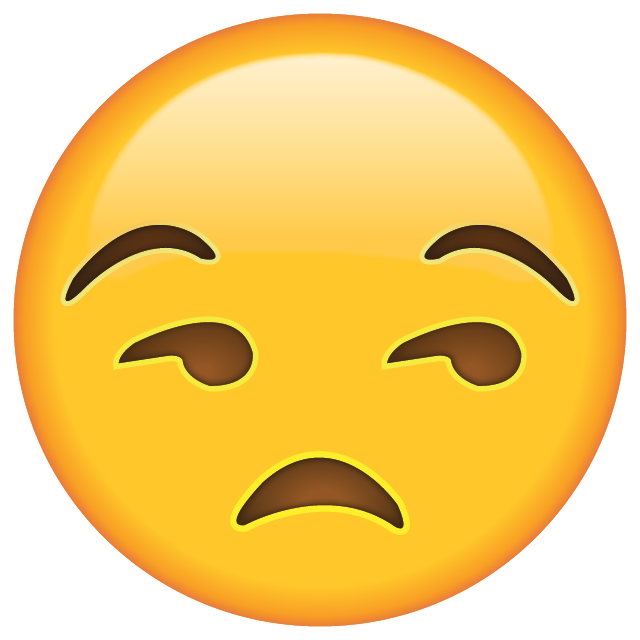}}
\subfloat[\label{}]{\includegraphics[width=0.08\linewidth]{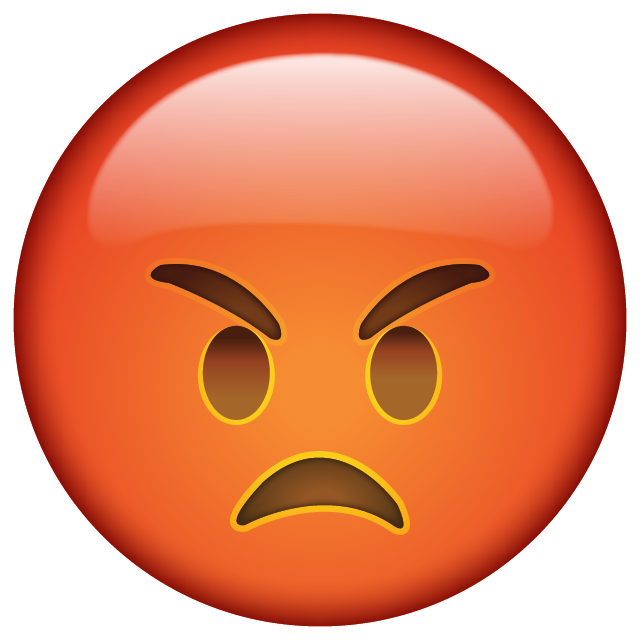}}
\subfloat[\label{}]{\includegraphics[width=0.08\linewidth]{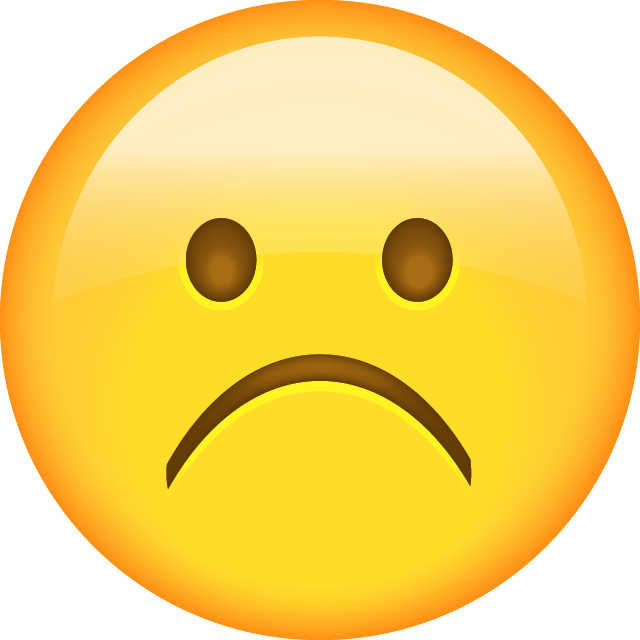}}
\subfloat[\label{}]{\includegraphics[width=0.08\linewidth]{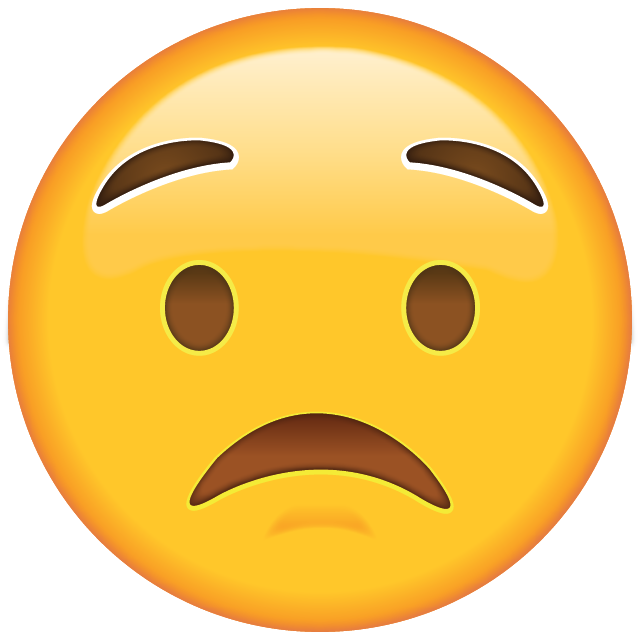}}
\caption{All emojis used to represent negative emotion. Their respective unicodes are: (a) U+1F620, (b) U+1F627, (c) U+1F61E, (d) U+1F628, (e) U+1F626, (f) U+1F623, (g) U+1F614, (h) U+1F629, (i) U+1F612, (j) U+1F621, (k) U+2639, and (l) U+1F61F.}\label{fig:negative_emojis}
\end{figure}

After collecting the tweets, the next step was to split them into positive and negative classes. In order to do so, we used lists of emojis and emoticons selected according to the sentiment conveyed by them. Therefore, the polarity of a tweet is determined by the presence of emojis and emoticons in it -- if it only contains positive ones (from the positive list), its polarity is assigned as positive. If a tweet contains both positive and negative elements, it is discarded since it is likely to be ambiguous. Following this idea, we used the same list of emoticons used by Go et al.~\cite{go2009twitter}, which is presented in Table~\ref{tab:emoticons}. Go et al.\cite{go2009twitter} did not use emojis, but these graphic symbols are also employed to convey ideas and sentiments~\cite{novak2015sentiment}. In contrast to the small set of emoticons, there are hundreds of possible emojis. Therefore, we selected a representative list with positive and negative emojis. All the emojis conveying positive emotion are presented in Fig.~\ref{fig:positive_emojis}. Fig.~\ref{fig:negative_emojis} illustrates the selected ones with negative emotion.


After filtering the tweets by the aforementioned criteria, we obtained a labeled corpus comprising 554,623 positive tweets and 425,444 negative ones. This corpus was used to train the machine learning methods. It is important to highlight that emojis and emoticons were removed from the tweets in the final corpus, so that their presence as a sentiment indicator is not learned by the models.

In addition to the filtering process, some preprocessing steps were performed to improve the corpus quality. Details about the preprocessing steps are given in the Supplementary Material, Section A. After these steps, tweets containing less than 4 tokens were discarded from the corpus. The complete framework (tweets collection, filtering and preprocessing methods) along with all experimental evaluation will be made available~\footnote{\url{https://github.com/edilsonacjr/pelesent}}.

\begin{table}[]
\centering
\caption{All emoticons used to represent emotion.}
\label{tab:emoticons}
\begin{tabular}{lcl}
\toprule
\textbf{Positive} & & \textbf{Negative} \\ 
\midrule
:) \quad :-) \quad :D \quad =)           &  & :( \quad :-(                \\
\bottomrule
\end{tabular}
\end{table}

\section{Experimental Evaluation}\label{sec:experiment}

In order to evaluate the quality of the corpus built using distant supervision, we trained state-of-the-art methods for polarity classification and applied the learned models to 5 well known manually annotated sentiment corpora. In the following, we present these corpora along with the message polarity classification methods, and finally, the obtained results.

\subsection{Corpora}

Sentiment classifiers are usually trained on manually annotated corpora. Because sentiments may be expressed differently in different domains~\cite{pan2010cross}, it is common to create domain-specific corpus. Since we intend to create a robust and generic corpus that is not domain-specific, we selected 5 corpora for evaluation, 2 being from the same domain (Twitter) and 3 from a different domain (product reviews). Below, we present the corpora that were used.

\paragraph{Brazilian Presidential Election~\cite{silva2011effective}} 
This dataset is formed by tweets about the Brazilian presidential election run in 2010. The corpus is divided in two parts, one referencing Dilma Rousseff~(\textit{BPE-Dilma}) and the other José Serra~(\textit{BPE-Serra}), both being the most popular candidates in the election. The corpora were manually annotated in positive and negative, and used to evaluate stream based sentiment analysis systems.

\paragraph{Buscap\'e~\cite{hartmann2014large}}
This dataset is formed by product reviews extracted from Buscapé website\footnote{\url{http://www.buscape.com.br/}}. The documents were automatically labeled based on two informations given by the users. The first~(\textit{Buscape-1}) is based on a recommendation tag while the second~(\textit{Buscape-2}) is based on a 5-star scale~(1-2 stars for negative and 4-5 stars for positive). Both corpora are balanced between the two classes, even though there is a notable difference on their sizes, possibly due to the low use of the recommendation tag. 

\paragraph{Mercado Livre~\cite{avanccoimproving}}
Similar to the Buscapé dataset, this corpus is formed by product reviews from the online marketplace Mercado Livre\footnote{\url{http://www.mercadolivre.com.br/}}. The corpus was also automatically annotated based on a 5-star scale given by the authors of the reviews. The dataset is balanced between the positive and negative classes. 

Table~\ref{tab:datasets} presents a summary of the corpora. 

\begin{table}[h]
\caption{\label{tab:datasets}Datasets used in the evaluation of the system.}
\begin{center}
\begin{tabular}{lrrr}
\toprule
\bf Dataset & \bf Total & \bf Positive & \bf Negative \\ 
\midrule
BPE-Dilma 	& $66,640$ & $46,805$ & $19,835$ \\
BPE-Serra 	& $9,718$ & $1,371$ & $8,347$ \\
Buscape-1 	& $2,000$ & $1,000$ & $1,000$ \\
Buscape-2 	& $13,685$ & $6,873$ & $6,812$ \\
Mercado Livre 	& $43,318$ & $21,819$ & $21,499$ \\
\bottomrule
\end{tabular}
\end{center}
\end{table}

\subsection{Machine Learning Methods}

Machine learning has dominated the area of sentiment analysis, mostly because its high performance when manually annotated data is available. However, thanks to the great variety of methods, there is no consensus about which method is the best in this scenario. In the last editions of SemEval Sentiment Analysis Task, most of the best methods/systems used deep learning techniques~\cite{rosenthal2015semeval,nakov2016semeval}. In this work, the evaluated methods range from simple linear models for classification using vector space models to hybrid (machine learning and lexical-based) and Deep Learning methods. The idea was to thoroughly evaluate the quality of the corpus regardless of the technique being used for learning. Below, each method is briefly described.

\paragraph{Logistic Regression (LR)} Also known as logit regression, LR can be understood as a generalization of linear regression models to the binary classification scenario, where a sigmoid function outputs the
class probabilities~\cite{Murphy:2012}. In this paper, the logistic regression model predicts the class probabilities of a text, where the classes are "positive" and "negative". As input for this classifier, three text representations were used: (1) a bag-of-words model (LR+tfidf), where each document (tweet or review) is represented by its set of words weighted by \textit{tf-idf}~\cite{salton1989automatic}; (2) a word embeddings based model (LR+w2v), where each document is represented by the weighted average of the embedding vectors (generated by Word2Vec~\cite{mikolov2013distributed,mikolov2013efficient}) of the words that compose the document, the weights are defined by \textit{tf-idf}; (3) the Paragraph Vector model (LR+d2v), which uses a neural network to generate embeddings for words and documents simultaneously in an unsupervised manner. Only the vectorial representations of documents were used by the classifier.

\paragraph{Convolutional Neural Networks (CNNs)} With the popularity of deep learning, CNNs have been applied to many different contexts, including several NLP tasks~\cite{collobert2011nlp} and, more specifically, sentiment analysis~\cite{kalchbrenner2014convolutional,kim2014cnn,severyn2015unitn,deriu2016swisscheese}. Our CNN is similar to the architecture proposed by Kim~\cite{kim2014cnn}, which uses a single convolutional layer. In this architecture, the network receives as input a matrix 
representing the document, and each word in the document is represented by a dense continuous vector. The output of the network is the probability of a document being negative or positive.

\paragraph{Recurrent Convolutional Neural Networks (RCNNs)} This deep neural architecture uses both convolutional and recurrent layers. Recently explored by many works in NLP~\cite{kalchbrenner2013recurrent,treviso2017sentence,lai2015recurrent}, this architecture has been successfully applied to sentiment analysis~\cite{lai2015recurrent,nakov2016semeval}.  Our architecture consists of a slight modification of the one used by Treviso et al.\cite{treviso2017sentence}, where the final layer returns the probability for the whole document, indicating a positive/negative polarity. Using this combination of convolutional and recurrent layers, we explored the principle that nearby words have a greater influence in the classification, while distant words may also have some impact. 

\paragraph{Hybrid} This method is a combination of two classifiers previously used for sentiment classification in cross-domain corpora~\cite{avanccoimproving} and follows the same setting introduced by Avan{\c{c}}o\cite{avancco2015normalizaccao}. The method consists of a SVM classifier combined with a lexical-based approach. The documents are represented by arrays of features including a binary bag-of-words (presence/absence of terms), emoticons, sentiment words and POS tags. Documents located near the separation hyperplane~(in a threshold assumed as $0.5$) learned by the SVM are considered to be uncertain. Those documents are then classified with a lexical-based approach, that uses linguistic rules for polarity classification in Portuguese. 

For all methods, well-known machine learning libraries were used, such as Scikit-learn~\cite{pedregosa2011scikit} and Keras~\cite{chollet2015keras}. Particularities such as parameters, details about the architecture, initializations and others can be found in the Supplementary Material, Section B.

\subsection{Results and Discussion}\label{sec:results}

To evaluate and compare the methods in each corpus, F1 score (macro-averaged), recall (macro-averaged) and accuracy were chosen, mostly because of their traditional use in sentiment analysis~\cite{rosenthal2015semeval,nakov2016semeval}. 

The main results are shown in Table~\ref{tab:scores}. Along with the results of each polarity classification method, we present the state-of-the-art (SotA) result reported for each corpus. Because the BPE corpora were conceived for a different context, there are no SotA reported results for those corpora. We also ranked each evaluated method by its F1 score.

The differences between the best method (in bold) and the SotA vary between $9.69\%$ and $12.24\%$, very competitive results given the fact that all SotA reported results were obtained by a 10-fold cross validation scheme and our methods used a corpus from a different domain for training. Of all the methods, the Hybrid was the one that had the best performance in the corpora of product reviews. Such a result was due to the regularity of the language in this type of corpus, which makes lexical approaches highly effective. However, in domains such as Twitter, errors, abbreviations and slangs are very common, which decreases the effectiveness of lexical-based approaches. This effect can be seen in the BPE-Dilma corpus.

An important aspect of Sentiment Analysis is the sensitivity of its methods to elements such as domain and temporality. In our evaluation, both were present in the selected corpora, which demonstrates the robustness of the constructed corpus and its resilience to temporality and the non-regularity of the language.

Regarding the deep learning methods (CNN and RCNN), both presented high rankings in almost all corpora. However, there was no huge difference between deep and shallow methods (logistic regression + document representation), indicating that large datasets decrease the performance difference between methods from different natures (even between simple and complex methods), a result commonly found in the big data era~\cite{halevy2009unreasonable}.

\begin{table}[h]
\caption{\label{tab:scores} Results obtained by each method trained on the distant supervision corpus.}
\begin{center}
\begin{tabular}{llccc}
\toprule
\bf Dataset & \bf Method & \bf $\mathbf{F_1}$ score & \bf Recall & \bf Accuracy\\ 
\midrule
 \multirow{5}{*}{BPE-Dilma} 
 & LR + w2v		&  $0.5739$\textsuperscript{5}  &  $0.6037$ &   $0.5952$ \\
 & LR + tfidf	&\bf 0.6477\textsuperscript{1}  &  $0.6443$ &\bf 0.7128 \\
 & LR + d2v     &  $0.6135$\textsuperscript{4}  &  $0.6071$ &   $0.7256$ \\
 & CNN			&   $0.6337$\textsuperscript{3} &  $0.6295$ &   $0.7051$ \\
 & RCNN 		&   $0.6444$\textsuperscript{2} &\bf 0.6586  &   $0.6816$ \\
 & Hybrid 		&   $0.5249$\textsuperscript{6} &  $0.5855$ &   $0.5295$ \\
 \cmidrule{2-5}
 & SotA			&  $-$ & $-$ & $-$ \\
\midrule

\multirow{5}{*}{BPE-Serra} 
 & LR + w2v		&   $0.3515$\textsuperscript{6} &   $0.4398$ &   $0.3915$ \\
 & LR + tfidf	&   $0.4110$\textsuperscript{5} &   $0.5546$ &   $0.4475$ \\
 & LR + d2v    	&   $0.5055$\textsuperscript{3} &   $0.6028$ &   $0.5915$ \\
 & CNN			&   $0.4240$\textsuperscript{4} &   $0.5929$ &   $0.4558$ \\
 & RCNN			&   $0.5286$\textsuperscript{2} &   $0.5975$ &   $0.6426$ \\
 & Hybrid		&\bf 0.5745\textsuperscript{1}  &\bf 0.6073  &\bf  0.7344 \\
 \cmidrule{2-5}
 & SotA			&  $-$ & $-$ & $-$ \\
\midrule

\multirow{5}{*}{Buscape-1} 
 & LR + w2v		&   $0.7232$\textsuperscript{4} &   $0.7250$ &   $0.7250$ \\
 & LR + tfidf	&   $0.7469$\textsuperscript{3} &   $0.7480$ &   $0.7480$ \\
 & LR + d2v    	&   $0.6427$\textsuperscript{6} &   $0.6465$ &   $0.6465$ \\
 & CNN			&   $0.6713$\textsuperscript{5} &   $0.6870$ &   $0.6870$ \\
 & RCNN			&   $0.7654$\textsuperscript{2} &   $0.7654$ &   $0.7654$ \\
 & Hybrid		&\bf 0.7668\textsuperscript{1}  &\bf  0.7695 &\bf  0.7695 \\
 \cmidrule{2-5}
 & SotA			&  $0.8892$ & $-$ & $0.8894$ \\
\midrule

\multirow{5}{*}{Buscape-2} 
 & LR + w2v		&   $0.6814$\textsuperscript{6} &   $0.6903$ &   $0.6910$ \\
 & LR + tfidf	&   $0.7725$\textsuperscript{2} &   $0.7738$ &   $0.7742$ \\
 & LR + d2v    	&   $0.7017$\textsuperscript{5} &   $0.7027$ &   $0.7030$ \\
 & CNN			&   $0.7048$\textsuperscript{4} &   $0.7115$ &   $0.7122$ \\
 & RCNN			&   $0.7656$\textsuperscript{3} &   $0.7658$ &   $0.7657$ \\
 & Hybrid		&\bf 0.7917\textsuperscript{1}  &\bf 0.7930  &\bf 0.7934  \\
 \cmidrule{2-5}
 & SotA			&  $0.8935$ & $-$ & $0.8935$ \\
\midrule

\multirow{5}{*}{Mercado Livre} 
 & LR + w2v		&   $0.6861$\textsuperscript{6} &   $0.7048$ &   $0.7066$ \\
 & LR + tfidf	&   $0.8328$\textsuperscript{3} &   $0.8329$ &   $0.8328$ \\
 & LR + d2v     &   $0.8089$\textsuperscript{4} &   $0.8093$ &   $0.8097$ \\
 & CNN			&   $0.7745$\textsuperscript{5} &   $0.7800$ &   $0.7813$ \\
 & RCNN			&   $0.8561$\textsuperscript{2} &   $0.8561$ &   $0.8563$ \\
 & Hybrid		&\bf  0.8614\textsuperscript{1} &\bf  0.8614 &\bf 0.8614  \\
 \cmidrule{2-5}
 & SotA			&  $0.9583$ & $-$ & $0.9583$ \\
\bottomrule
\end{tabular}
\end{center}
\end{table}

\section{Conclusion and Future Work}\label{sec:conclusion}

In recent years, the polarity classification task has drawn the attention of the scientific community, mainly due to its direct application in scenarios such as social media content and product reviews. Even though machine learning methods present themselves as high performance alternatives, they suffer from the need of a large amount of data during their training phases. In this paper, we adapted a distant supervision approach to build a large sentiment corpus for Portuguese. State-of-the-art methods were trained on this corpus and applied to 5 selected corpora, from same domain and different domain (cross-domain). Competitive results were obtained for all methods, although our best results did not outperform the best ones reported for the same corpora.

As future works, we intend to explore ways to improve the quality of the distant supervision corpus by applying techniques to remove outliers and tweets that do not convey any sentiment or convey the wrong sentiment. We also intend to modify this framework to make it able to represent the neutral class.





\section*{Acknowledgment}

E.A.C.J. acknowledges financial support from Google (Google Research Awards in Latin America grant) and CAPES (Coordination for the Improvement of Higher Education Personnel). V.Q.M. acknowledges financial support from FAPESP (grant no. 15/05676-8). L.B.S. acknowledges financial support from Google (Google Research Awards in Latin America grant) and CNPq (National Council for Scientific and Technological Development, Brazil). T.F.C.B., M.V.T., and H.B.B. acknowledge financial support from CNPq. In part of this work a GPU donated by NVIDIA Corporation was used.

\bibliographystyle{IEEEtran}
\bibliography{biblio}

\begin{thebibliography}{10}
\providecommand{\url}[1]{#1}
\csname url@samestyle\endcsname
\providecommand{\newblock}{\relax}
\providecommand{\bibinfo}[2]{#2}
\providecommand{\BIBentrySTDinterwordspacing}{\spaceskip=0pt\relax}
\providecommand{\BIBentryALTinterwordstretchfactor}{4}
\providecommand{\BIBentryALTinterwordspacing}{\spaceskip=\fontdimen2\font plus
\BIBentryALTinterwordstretchfactor\fontdimen3\font minus
  \fontdimen4\font\relax}
\providecommand{\BIBforeignlanguage}[2]{{%
\expandafter\ifx\csname l@#1\endcsname\relax
\typeout{** WARNING: IEEEtran.bst: No hyphenation pattern has been}%
\typeout{** loaded for the language `#1'. Using the pattern for}%
\typeout{** the default language instead.}%
\else
\language=\csname l@#1\endcsname
\fi
#2}}
\providecommand{\BIBdecl}{\relax}
\BIBdecl

\bibitem{pang2008opinion}
B.~Pang, L.~Lee \emph{et~al.}, ``Opinion mining and sentiment analysis,''
  \emph{Foundations and Trends in Information Retrieval}, vol.~2, no. 1--2, pp.
  1--135, 2008.

\bibitem{rosenthal2015semeval}
S.~Rosenthal, P.~Nakov, S.~Kiritchenko, S.~M. Mohammad, A.~Ritter, and
  V.~Stoyanov, ``Semeval-2015 task 10: Sentiment analysis in twitter,'' in
  \emph{Proceedings of the 9th international workshop on semantic evaluation
  (SemEval 2015)}, 2015, pp. 451--463.

\bibitem{nakov2016semeval}
P.~Nakov, A.~Ritter, S.~Rosenthal, F.~Sebastiani, and V.~Stoyanov,
  ``Semeval-2016 task 4: Sentiment analysis in twitter,'' \emph{Proceedings of
  the 10th international workshop on semantic evaluation (SemEval 2016)}, pp.
  1--18, 2016.

\bibitem{pan2010cross}
S.~J. Pan, X.~Ni, J.-T. Sun, Q.~Yang, and Z.~Chen, ``Cross-domain sentiment
  classification via spectral feature alignment,'' in \emph{Proceedings of the
  19th international conference on World wide web}.\hskip 1em plus 0.5em minus
  0.4em\relax ACM, 2010, pp. 751--760.

\bibitem{go2009twitter}
A.~Go, R.~Bhayani, and L.~Huang, ``Twitter sentiment classification using
  distant supervision,'' \emph{CS224N Project Report, Stanford}, vol.~1,
  no.~12, 2009.

\bibitem{souza2011construction}
M.~Souza, R.~Vieira, D.~Busetti, R.~Chishman, I.~M. Alves \emph{et~al.},
  ``Construction of a portuguese opinion lexicon from multiple resources,'' in
  \emph{8th Brazilian Symposium in Information and Human Language Technology},
  2011, pp. 59--66.

\bibitem{BalageFilho2013}
P.~P. Balage~Filho, T.~Pardo, and S.~Alu\'{\i}sio, ``An evaluation of the
  {Brazilian Portuguese LIWC} dictionary for sentiment analysis,'' in
  \emph{Proceedings of the 9th Brazilian Symposium in Information and Human
  Language Technology (STIL)}, S.~M. Alu\'{\i}sio and V.~D. Feltrim, Eds.\hskip
  1em plus 0.5em minus 0.4em\relax Fortaleza-CE, Brazil: Sociedade Brasileira
  de Computa\c{c}\~{a}o, 21--23~Oct. 2013, pp. 215--219.

\bibitem{avancco2014lexicon}
L.~V. Avan{\c{c}}o and M.~d. G.~V. Nunes, ``Lexicon-based sentiment analysis
  for reviews of products in brazilian portuguese,'' in \emph{Intelligent
  Systems (BRACIS), 2014 Brazilian Conference on}.\hskip 1em plus 0.5em minus
  0.4em\relax IEEE, 2014, pp. 277--281.

\bibitem{avanccoimproving}
L.~V. Avan{\c{c}}o, H.~B. Brum, and M.~G. Nunes, ``Improving opinion
  classifiers by combining different methods and resources,'' in \emph{XIII
  Encontro Nacional de Intelig\^encia Artificial e Computacional (ENIAC)},
  2016, pp. 25--36.

\bibitem{mikolov2013efficient}
T.~Mikolov, K.~Chen, G.~Corrado, and J.~Dean, ``Efficient estimation of word
  representations in vector space,'' \emph{arXiv preprint arXiv:1301.3781},
  2013.

\bibitem{mikolov2013distributed}
T.~Mikolov, I.~Sutskever, K.~Chen, G.~S. Corrado, and J.~Dean, ``Distributed
  representations of words and phrases and their compositionality,'' in
  \emph{Advances in neural information processing systems}, 2013, pp.
  3111--3119.

\bibitem{zhou2016ecnu}
Y.~Zhou, Z.~Zhang, and M.~Lan, ``Ecnu at semeval-2016 task 4: An empirical
  investigation of traditional nlp features and word embedding features for
  sentence-level and topic-level sentiment analysis in twitter,''
  \emph{Proceedings of SemEval}, pp. 256--261, 2016.

\bibitem{correa2017nilc}
E.~A. Corr{\^e}a~Jr, V.~Q. Marinho, and L.~B.~d. Santos, ``Nilc-usp at
  semeval-2017 task 4: A multi-view ensemble for twitter sentiment analysis,''
  \emph{Proceedings of the 11th International Workshop on Semantic Evaluation
  (SemEval'17)}, 2017.

\bibitem{mikolovdoc2vec}
Q.~Le and T.~Mikolov, ``Distributed representations of sentences and
  documents,'' in \emph{Proceedings of the 31st International Conference on
  Machine Learning (ICML-14)}, T.~Jebara and E.~P. Xing, Eds., 2014, pp.
  1188--1196.

\bibitem{socher2013recursive}
R.~Socher, A.~Perelygin, J.~Y. Wu, J.~Chuang, C.~D. Manning, A.~Y. Ng, C.~Potts
  \emph{et~al.}, ``Recursive deep models for semantic compositionality over a
  sentiment treebank,'' in \emph{Proceedings of the conference on empirical
  methods in natural language processing (EMNLP)}, vol. 1631.\hskip 1em plus
  0.5em minus 0.4em\relax Citeseer, 2013, p. 1642.

\bibitem{kalchbrenner2014convolutional}
\BIBentryALTinterwordspacing
N.~Kalchbrenner, E.~Grefenstette, and P.~Blunsom, ``A convolutional neural
  network for modelling sentences,'' in \emph{Proceedings of the 52nd Annual
  Meeting of the Association for Computational Linguistics (Volume 1: Long
  Papers)}.\hskip 1em plus 0.5em minus 0.4em\relax Baltimore, Maryland:
  Association for Computational Linguistics, June 2014, pp. 655--665. [Online].
  Available: \url{http://www.aclweb.org/anthology/P14-1062}
\BIBentrySTDinterwordspacing

\bibitem{kim2014cnn}
Y.~Kim, ``Convolutional neural networks for sentence classification,'' in
  \emph{Proceedings of the 2014 Conference on Empirical Methods in Natural
  Language Processing (EMNLP)}.\hskip 1em plus 0.5em minus 0.4em\relax Doha,
  Qatar: Association for Computational Linguistics, October 2014, pp.
  1746--1751.

\bibitem{severyn2015unitn}
A.~Severyn and A.~Moschitti, ``Unitn: Training deep convolutional neural
  network for twitter sentiment classification,'' in \emph{Proceedings of the
  9th International Workshop on Semantic Evaluation (SemEval 2015), Association
  for Computational Linguistics, Denver, Colorado}, 2015, pp. 464--469.

\bibitem{deriu2016swisscheese}
J.~Deriu, M.~Gonzenbach, F.~Uzdilli, A.~Lucchi, V.~De~Luca, and M.~Jaggi,
  ``Swisscheese at semeval-2016 task 4: Sentiment classification using an
  ensemble of convolutional neural networks with distant supervision,''
  \emph{Proceedings of SemEval}, pp. 1124--1128, 2016.

\bibitem{read2005using}
J.~Read, ``Using emoticons to reduce dependency in machine learning techniques
  for sentiment classification,'' in \emph{Proceedings of the ACL student
  research workshop}.\hskip 1em plus 0.5em minus 0.4em\relax Association for
  Computational Linguistics, 2005, pp. 43--48.

\bibitem{pak2010twitter}
A.~Pak and P.~Paroubek, ``Twitter as a corpus for sentiment analysis and
  opinion mining.'' in \emph{LREc}, vol.~10, no. 2010, 2010.

\bibitem{davidov2010enhanced}
D.~Davidov, O.~Tsur, and A.~Rappoport, ``Enhanced sentiment learning using
  twitter hashtags and smileys,'' in \emph{Proceedings of the 23rd
  international conference on computational linguistics: posters}.\hskip 1em
  plus 0.5em minus 0.4em\relax Association for Computational Linguistics, 2010,
  pp. 241--249.

\bibitem{kouloumpis2011twitter}
E.~Kouloumpis, T.~Wilson, and J.~Moore, ``Twitter sentiment analysis: The good
  the bad and the omg,'' in \emph{In The International Association for the
  Advancement of Artificial Intelligence Conference on Weblogs and
  Social}.\hskip 1em plus 0.5em minus 0.4em\relax Association for the
  Advancement of Artificial Intelligence, 2011.

\bibitem{novak2015sentiment}
P.~K. Novak, J.~Smailovi{\'c}, B.~Sluban, and I.~Mozeti{\v{c}}, ``Sentiment of
  emojis,'' \emph{PloS one}, vol.~10, no.~12, p. e0144296, 2015.

\bibitem{silva2011effective}
I.~S. Silva, J.~Gomide, A.~Veloso, W.~Meira~Jr, and R.~Ferreira, ``Effective
  sentiment stream analysis with self-augmenting training and demand-driven
  projection,'' in \emph{Proceedings of the 34th international ACM SIGIR
  conference on Research and development in Information Retrieval}.\hskip 1em
  plus 0.5em minus 0.4em\relax ACM, 2011, pp. 475--484.

\bibitem{hartmann2014large}
N.~S. Hartmann, L.~V. Avan{\c{c}}o, P.~P. Balage~Filho, M.~S. Duran, M.~D.
  G.~V. Nunes, T.~A.~S. Pardo, S.~M. Aluisio \emph{et~al.}, ``A large corpus of
  product reviews in portuguese: Tackling out-of-vocabulary words,'' in
  \emph{International Conference on Language Resources and Evaluation,
  9th.}\hskip 1em plus 0.5em minus 0.4em\relax European Language Resources
  Association-ELRA, 2014.

\bibitem{Murphy:2012}
K.~P. Murphy, \emph{Machine Learning: A Probabilistic Perspective}.\hskip 1em
  plus 0.5em minus 0.4em\relax The MIT Press, 2012.

\bibitem{salton1989automatic}
G.~Salton, ``Automatic text processing: The transformation, analysis, and
  retrieval of,'' \emph{Reading: Addison-Wesley}, 1989.

\bibitem{collobert2011nlp}
R.~Collobert, J.~Weston, L.~Bottou, M.~Karlen, K.~Kavukcuoglu, and P.~Kuksa,
  ``Natural language processing (almost) from scratch,'' \emph{Journal of
  Machine Learning Research}, vol.~12, no. Aug, pp. 2493--2537, 2011.

\bibitem{kalchbrenner2013recurrent}
N.~Kalchbrenner and P.~Blunsom, ``Recurrent convolutional neural networks for
  discourse compositionality,'' \emph{arXiv preprint arXiv:1306.3584}, 2013.

\bibitem{treviso2017sentence}
M.~V. Treviso, C.~Shulby, and S.~M. Alu{\'\i}sio, ``Sentence segmentation in
  narrative transcripts from neuropsychological tests using recurrent
  convolutional neural networks,'' in \emph{Proceedings of the 15th Conference
  of the European Chapter of the Association for Computational Linguistics:
  Volume 1, Long Papers}, 2017, pp. 315--325.

\bibitem{lai2015recurrent}
S.~Lai, L.~Xu, K.~Liu, and J.~Zhao, ``Recurrent convolutional neural networks
  for text classification.'' in \emph{AAAI}, vol. 333, 2015, pp. 2267--2273.

\bibitem{avancco2015normalizaccao}
L.~V. Avan{\c{c}}o, ``Sobre normaliza{\c{c}}{\~a}o e classifica{\c{c}}{\~a}o de
  polaridade de textos opinativos na web,'' Master's thesis, Universidade de
  S{\~a}o Paulo, 2015.

\bibitem{pedregosa2011scikit}
F.~Pedregosa, G.~Varoquaux, A.~Gramfort, V.~Michel, B.~Thirion, O.~Grisel,
  M.~Blondel, P.~Prettenhofer, R.~Weiss, V.~Dubourg \emph{et~al.},
  ``Scikit-learn: Machine learning in python,'' \emph{Journal of Machine
  Learning Research}, vol.~12, no. Oct, pp. 2825--2830, 2011.

\bibitem{chollet2015keras}
F.~Chollet \emph{et~al.}, ``Keras,'' \url{https://github.com/fchollet/keras},
  2015.

\bibitem{halevy2009unreasonable}
A.~Halevy, P.~Norvig, and F.~Pereira, ``The unreasonable effectiveness of
  data,'' \emph{IEEE Intelligent Systems}, vol.~24, no.~2, pp. 8--12, 2009.

\bibitem{Art:Srivastava:2014:dropout}
N.~Srivastava, G.~E. Hinton, A.~Krizhevsky, I.~Sutskever, and R.~Salakhutdinov,
  ``Dropout: a simple way to prevent neural networks from overfitting.''
  \emph{Journal of Machine Learning Research}, vol.~15, no.~1, pp. 1929--1958,
  2014.

\bibitem{Art:Hochreiter:1997:lstm}
S.~Hochreiter and J.~Schmidhuber, ``Long short-term memory,'' \emph{Neural
  computation}, vol.~9, no.~8, pp. 1735--1780, 1997.

\end{thebibliography}


\section*{Supplementary Material}
\label{sec:supplemental}

\subsection{Preprocessing}
In order to properly tokenize and preprocess the tweets, the following steps were performed:
\begin{itemize}
\item Punctuation marks forming an emoticon were considered as a single token (e.g. \emph{:-(} and \emph{;)} )
\item Sequences of consecutive emojis were split so that each emoji formed a single token 
\item Additional punctuation marks (not forming any emoticon) were removed
\item Usernames (an \emph{@} symbol followed by up to 15 characters) were replaced by the tag \emph{USERNAME}
\item Hashtags (a \emph{\#} symbol followed by any sequence of characters) were replaced by the tag \emph{HASHTAG}
\item URLs were completely replaced by the tag \emph{URL}
\item Numbers, including dates, telephone numbers and currency values were replaced by the tag \emph{NUMBER}
\item Subsequent character repetitions were limited to 3 -- i.e., all sequences of the same character were trimmed to fit the limit of 3
\end{itemize}

The following tweet is used to illustrate the preprocessing steps:

\textbf{Original:} \emph{hj tenho aula de manhã, tarde e noite. das 8h ate 19h :(( \#cansado}

\textbf{Preprocessed:} \emph{hj tenho aula de manhã tarde e noite das NUMBER ate NUMBER :(( HASHTAG}

\subsection{Details about the machine learning methods}

\paragraph{Logistic Regression} There is no additional information about this classifier.

\begin{figure}[!htb]
  \center
  \includegraphics[width=0.5\textwidth]{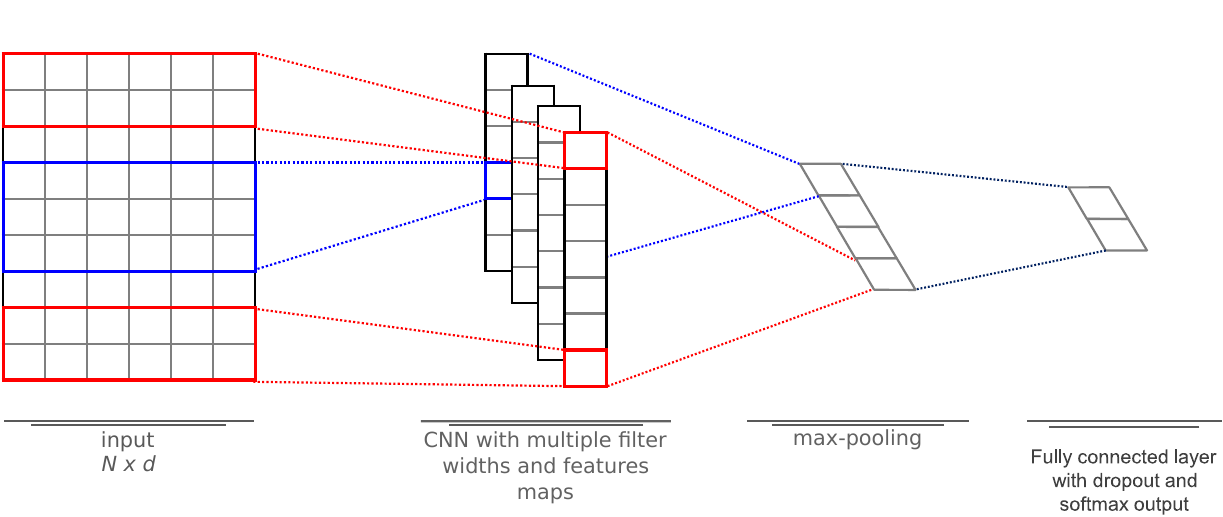}
  \caption{CNN architecture adapted from~\cite{kim2014cnn}.}
  \label{fig:architecture_cnn}
\end{figure}

\paragraph{Convolutional Neural Networks} 
The complete architecture is presented in Figure~\ref{fig:architecture_cnn}, where the input layer is a matrix composed by $n$ input words, and each word is a $d$ dimensionality real vector. The convolutional layer receives these vectors as input and it is responsible for the automatic extraction of $n_f$ features depending on a sliding window of length $h = \{3,4,5\}$. The output from the convolutional layer is then passed to the max-overtime pooling layer, and the new extracted features are concatenated. This results in a large dimensional
vector that is passed to a fully connected layer, where the softmax operation \cite{Murphy:2012} is applied, returning the probability of a document being negative or positive. In the penultimate layer, we employed dropout with a constraint on l2-norms of the weight vectors to reduce the chance of overfitting \cite{Art:Srivastava:2014:dropout}. 

\begin{figure}[!htb]
  \centering
  \includegraphics[width=0.5\textwidth]{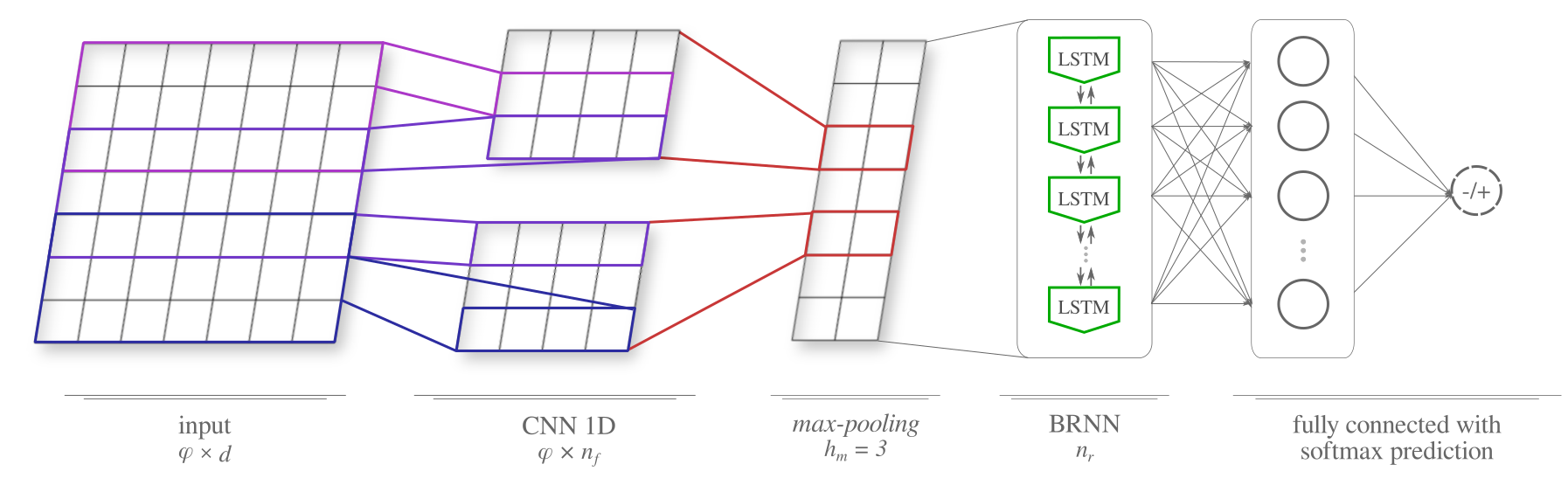}
  \caption{RCNN architecture adapted from~\cite{treviso2017sentence}. }
  \label{fig:architecture_rcnn}
\end{figure}

\paragraph{Recurrent Convolutional Neural Networks}
The complete architecture is illustrated in Figure~\ref{fig:architecture_rcnn}. The architecture is composed by an input layer that has $\varphi$ input features, and each feature has a dimensionality of $d$. The convolutional layer is responsible for the automatic extraction of $n_f$ new features depending on $3$ neighboring words. Then, a max-pooling operation is applied over time, looking at a region of $h_m=3$ elements to find the most significant features. The new extracted features are fed into a recurrent bidirectional layer which has $n_f$ units known as Long Short-Term Memory \cite{Art:Hochreiter:1997:lstm}, which are able to learn over long dependencies between words. Finally, the last recurrent state output is passed to a totally connected layer, where the softmax operation \cite{Murphy:2012} is calculated, giving the probability of a document being negative or positive. Between these two layers, dropout is used to reduce the chance of overfitting \cite{Art:Srivastava:2014:dropout}.

For both neural network models (CNN and RCNN), we employed the early stopping strategy ($p=3$) to avoid overfitting, i.e. the training phase finishes when the validation loss has stopped improving. Other experimental settings for CNN and RCNN (number of epochs, batch size, learning rate, etc.) can be seen in their original papers~\cite{kim2014cnn,treviso2017sentence}, respectively.

\paragraph{Hybrid}
The two methods combined to classify a document in polarity classes are described below:
\begin{itemize}
\item SVM classifier: The SVM employed uses a RBF Kernel~(gamma $=1/$n\_features), $C$ defined as $0.25$, and L1 penalty for regularization.

\item Lexical-based classifier: Each word present in a sentiment lexicon receives a value according to its polarity. Positive words are valued as $1$ and negative ones as $-1$. The presence of an intensification word~(e.g. \textit{muito}, \textit{demais}) in a window around the word multiplies its value by $3$. The presence of a downtoner divides the current value by $3$. A negation multiplies the value of a word by $-1$, inverting its polarity. Whenever a negation is in the same window as an intensification, it becomes a downtoner~(e.g. \textit{não muito}), the same occurs with a downtoner~(\textit{não pouco}). The polarity values are then summed up to determine the document polarity.
\end{itemize}

\end{document}